\documentclass[]{lab}

\usepackage[utf8]{inputenc}
\usepackage[T1]{fontenc}
\usepackage{geometry}
\usepackage{amsmath,amssymb}  %
\usepackage{charter}

\usepackage[toc,page,header]{appendix}

\usepackage{cleveref} 
\usepackage{subcaption}
\usepackage{booktabs}
\usepackage{graphicx}
\usepackage{xcolor}
\usepackage{CJKutf8}
\usetikzlibrary{patterns}
\usepackage{float}
\usepackage{placeins}
\usepackage{caption}
\usepackage{bm}

\usepackage{soul} %
\usepackage{algorithm}      %
\usepackage{algpseudocode}  %
\usepackage{parskip}

\definecolor{cvprblue}{rgb}{0.21,0.49,0.74}
\definecolor{fallbackgreen}{rgb}{130, 180, 102}
\definecolor{stopred}{rgb}{251, 225, 224}

\ifdefined\final
\usepackage[disable]{todonotes}
\else
\usepackage[textsize=tiny]{todonotes}
\fi

\usepackage{natbib}
\usepackage{latexsym}

\usepackage{url}
\usepackage{amssymb}
\usepackage[utf8]{inputenc}
\usepackage{microtype}
\usepackage{booktabs}
\usepackage{pifont} 
\usepackage{multirow}
\usepackage{makecell}
\usepackage{paralist}
\usepackage{xspace}
\usepackage{color}
\usepackage{xcolor}
\usepackage{colortbl}
\usepackage{adjustbox}
\usepackage{hyperref} 
\usepackage[edges]{forest}
\usepackage{tikz} 
\usepackage{caption}
\usepackage{amsfonts}

\hypersetup{
    colorlinks,
    linkcolor={blue!80!black},
    citecolor={blue!80!black},
}
\tikzset{
    root/.style =             {align=center, text width=1cm, rounded corners=3pt, line width=0.3mm, fill=gray!10, draw=gray!80, font=\small},
    demographic/.style =         {align=center, text width=1.8cm, rounded corners=3pt, line width=0.3mm, fill=blue!10, draw=blue!80, font=\footnotesize},
    demographic_work/.style =    {align=center, text width=10cm, rounded corners=3pt, line width=0.3mm, fill=blue!10, draw=blue!0, font=\footnotesize},
    character/.style =         {align=center, text width=1.8cm, rounded corners=3pt, line width=0.3mm, fill=red!10, draw=red!80, font=\footnotesize},
    character_work/.style =    {align=center, text width=10cm, rounded corners=3pt, line width=0.3mm, fill=red!10, draw=red!0, font=\footnotesize},
    personalization/.style =           {align=center, text width=1.8cm, rounded corners=3pt, line width=0.3mm, fill=cyan!10, draw=cyan!80, font=\footnotesize},
    personalization_work/.style =      {align=center, text width=10cm, rounded corners=3pt, line width=0.3mm, fill=cyan!10, draw=cyan!0, font=\footnotesize},
    risk/.style =         {align=center, text width=1.8cm, rounded corners=3pt, line width=0.3mm, fill=orange!10, draw=orange!80, font=\footnotesize},
    risk_work/.style =    {align=center, text width=10cm, rounded corners=3pt, line width=0.3mm, fill=orange!10, draw=orange!0, font=\footnotesize},
}

\usepackage{CJK}

\newtcolorbox{promptbox}[1][]{
  enhanced,
  breakable,
  colback=promptboxlightgray,
  colframe=promptboxblue!30,
  arc=8pt,
  boxrule=0.5pt,
  left=12pt,
  right=12pt,
  top=8pt,
  bottom=8pt,
  fonttitle=\bfseries,
  fontupper=\linespread{1.2}\selectfont,
  title=#1
}

\title{World Action Models in Real Time: An Empirical Study of Smooth Execution via Asynchronous Deployment}

\author{Motubrain Team \\Project Page: \url{https://www.motubrain.cn/zh/}
\\GitHub: \url{https://github.com/shengshu-ai/Motubrain}
}

\begin{document}

\maketitle


\section{Introduction}
\label{sec:intro}

World Action Models---generative models that jointly reason over perception and action to predict robot motion---have demonstrated impressive dexterity across diverse manipulation tasks~\cite{bi2025motusunifiedlatentaction,fastwam2026,lingbotva2026,motubrain2026}. Building on Diffusion Policy~\cite{chi2023diffusion}, these models predict fixed-horizon action chunks via iterative denoising. A common deployment strategy is \textit{chunk execution}~\cite{zhao2023act}: predict a fixed-horizon sequence of actions and execute the whole sequence open-loop before re-querying the model. This amortizes inference cost but introduces significant deployment challenges.

World Action Models require iterative denoising over joint video-action sequences; combined with camera capture, network transmission, and command interpolation, the total end-to-end pipeline latency reaches the order of seconds. This has two compounding consequences: first, the robot must freeze or replay stale actions while awaiting each new prediction, producing visible discontinuities at every chunk boundary; second, the effective closed-loop rate is bounded by the inference frequency, leaving the system unresponsive to environmental changes---such as a suddenly appearing obstacle or a shifted object pose---during the entire inference window.

Asynchronous inference---triggering a new inference call before the current chunk is consumed---addresses both issues by overlapping compute with execution. The challenge is that consecutive chunks predicted from different observations disagree in their overlapping region; naively splicing them at the boundary produces the same or worse jitter. Prior work on Vision-Language-Action (VLA) model deployment~\cite{black2025rtc,black2024training,liu2025legato} proposes a range of strategies to handle this disagreement, which we organize into four families:

\begin{itemize}
  \item \textbf{Pure asynchronous switching}: execute a fixed prefix of each chunk and switch without any blending.
  \item \textbf{Guided diffusion}: steer the denoising trajectory of the incoming chunk toward consistency with the outgoing chunk at inference time (e.g., RTC~\cite{black2025rtc}).
  \item \textbf{Prefix-conditioned}: supply the committed portion of the prior chunk as a clean conditioning signal during \textit{training}, so the model learns to produce consistent continuations (e.g., Training-Time RTC~\cite{black2024training}).
  \item \textbf{Direct action weighting}: blend the denoised output actions of adjacent chunks via weighted interpolation.
\end{itemize}

\begin{table}[htbp]
\centering
\small
\caption{The four method families and their instantiations evaluated in this paper. \texttt{sync} is included as a non-async baseline.}
\label{tab:overview}
\begin{tabular}{llll}
\toprule
Family & Key mechanism & Methods & Retraining \\
\midrule
Sync baseline & Wait for inference; hard switch & \texttt{sync} & No \\
\midrule
Pure async switching & Switch at delay offset; no blending & \texttt{async} & No \\
Direct action weighting & Weighted blend of actions from adjacent chunks & \texttt{async+blend}, \texttt{simple} & No \\
Guided diffusion & Steer denoising velocity at inference time & \texttt{infer} & No \\
Prefix-conditioned & Condition generation on prior chunk prefix & \texttt{train} & Yes \\
\bottomrule
\end{tabular}
\end{table}

In this paper we implement and compare methods from all four families on our robot platform. Our contributions are:

\begin{enumerate}
  \item We show that temporal alignment---accurately associating each command frame with the correct observation timestamp---is a critical and often underappreciated engineering requirement; poor alignment causes persistent jitter that no blending algorithm can compensate for.
  \item We provide a unified comparison of the four method families using both offline trajectory metrics and online task completion score, time, and smoothness across multiple manipulation scenarios.
  \item We find that prefix-conditioned methods achieve the best overall precision--smoothness balance, while direct action weighting provides a smooth but precision-limited baseline, and velocity-guided inference fails to constrain the delay region on our platform.
\end{enumerate}

\section{Background and Related Work}

\subsection{Action Chunk Policies and World Action Models}

Diffusion Policy~\cite{chi2023diffusion} introduced the idea of predicting a fixed-horizon action chunk $\mathbf{a}_{t:t+H}$ conditioned on an observation $\mathbf{o}_t$ via iterative denoising. Subsequent work scaled this paradigm to vision-language-conditioned World Action Models~\cite{lingbotva2026,motubrain2026,dreamzero2026} capable of generalizing across diverse manipulation tasks. Chunk execution amortizes the high cost of denoising inference but commits the robot to a stale plan for up to $H$ steps, motivating the asynchronous deployment strategies studied in this work.

\subsection{Asynchronous Deployment Strategies}

Running a new inference call before the current chunk is consumed decouples execution frequency from inference frequency, but introduces inter-chunk disagreement in the overlapping region. Prior work proposes several strategies to resolve this disagreement.

\textbf{Pure asynchronous switching} simply executes a prefix of each chunk and switches at a fixed offset, accepting potential boundary discontinuities.

\textbf{Guided diffusion methods} address the disagreement at inference time by steering the denoising trajectory of the incoming chunk toward the outgoing chunk. RTC~\cite{black2025rtc} formulates this as an inpainting-style constraint on the denoising velocity field at each denoising step,   en utilisant le  prior chunk as a soft target to guide both the trajectory direction and the resulting actions in the overlap region.

\textbf{Prefix-conditioned methods} address the disagreement at training time by injecting the committed portion of the prior chunk as a clean conditioning signal, teaching the model to produce continuations consistent with prior commitments. Training-Time RTC~\cite{black2024training} is the representative method in this family.

\textbf{Direct action weighting} operates purely on the output action sequences, blending adjacent chunks via weighted interpolation without modifying the training objective.

Some methods combine multiple families. Legato~\cite{liu2025legato} combines prefix-conditioned training with direct action weighting, and reports improvements over RTC in smoothness and task completion time.


\begin{figure}[htbp]
  \centering
  \includegraphics[width=\linewidth]{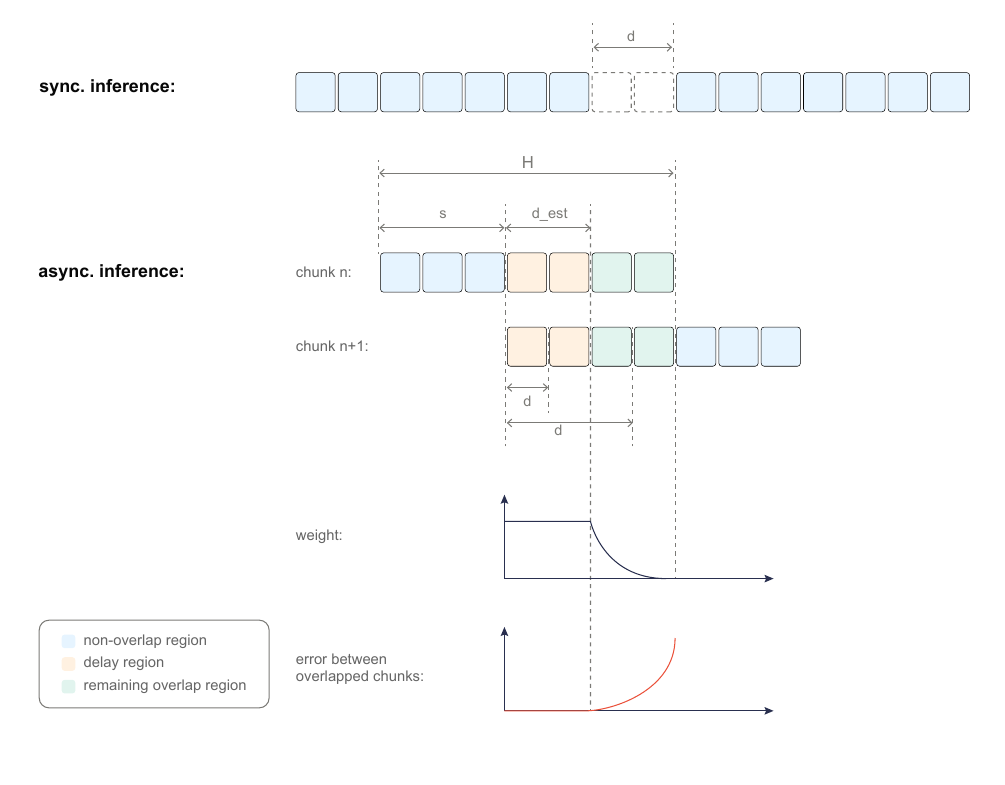}
  \caption{Timing diagram for synchronous (top) and asynchronous (bottom) chunk execution. In async mode, a chunk is divided into the delay region (yellow), remaining overlap (green), and non-overlap region (blue), where the first two form the overlap with the prior chunk.}
  \label{fig:timing}
\end{figure}
\section{Problem Formulation}

We consider a chunk-based policy that predicts a fixed-horizon action sequence of length $H$ at each inference call. Let chunk $n$ denote the action sequence $\mathbf{a}^{(n)} = [\mathbf{a}^{(n)}_0, \ldots, \mathbf{a}^{(n)}_{H-1}]$ predicted from observation $\mathbf{o}^{(n)}$. We define four key quantities (see Figure~\ref{fig:timing}):

\begin{itemize}
  \item $H$: prediction horizon (number of frames per chunk)
  \item $s$: execution horizon---the number of frames executed from chunk $n$ before the next inference call is triggered ($s < H$)
  \item $d$: true end-to-end pipeline delay in frames (camera capture $\to$ network $\to$ inference $\to$ command dispatch)
  \item $d_\text{est}$: estimated delay used by the deployment controller
\end{itemize}

\textbf{Synchronous execution} waits for inference to complete before switching chunks, leaving the robot idle or replaying the last action for $d$ frames at every chunk boundary (Figure~\ref{fig:timing}, top).

\textbf{Asynchronous execution} triggers inference after executing $s$ frames of chunk $n$, so that chunk $n{+}1$ is ready approximately when chunk $n$ has been executed for $s + d_\text{est}$ frames (Figure~\ref{fig:timing}, bottom). The two chunks overlap for $H - s$ frames in total. Within this overlap we distinguish two sub-regions:

\begin{itemize}
  \item \textbf{Delay region} (yellow, $d_\text{est}$ frames): the first $d_\text{est}$ frames of the overlap, corresponding to the pipeline delay. Chunk $n$ will execute these frames while chunk $n{+}1$ is being generated. Ideally, both chunks should produce identical predictions for this region, as these actions are already committed from chunk $n$ and serve as the known prefix for prefix-conditioned methods.
  \item \textbf{Remaining overlap} (green, $H - s - d_\text{est}$ frames): further frames predicted by both chunks. Since these frames lie further into the future, we expect inter-chunk disagreement to increase with prediction horizon.
\end{itemize}

The final $s$ frames of chunk $n{+}1$ have no counterpart in chunk $n$ and form the \textbf{non-overlap region} (blue).

The true delay $d$ fluctuates at runtime due to network jitter, GPU scheduling variance, and command-queue depth, making it difficult to estimate precisely. When $d > d_\text{est}$, the controller switches to chunk $n{+}1$ at a frame index that does not correspond to the robot's actual state, producing a hard position jump that no blending strategy can recover from. Conversely, when $d < d_\text{est}$, the switch happens slightly early and the impact is generally mild.

\section{Methods}
\label{sec:methods}

\begin{figure*}[htbp]
    \centering
    \includegraphics[width=\textwidth]{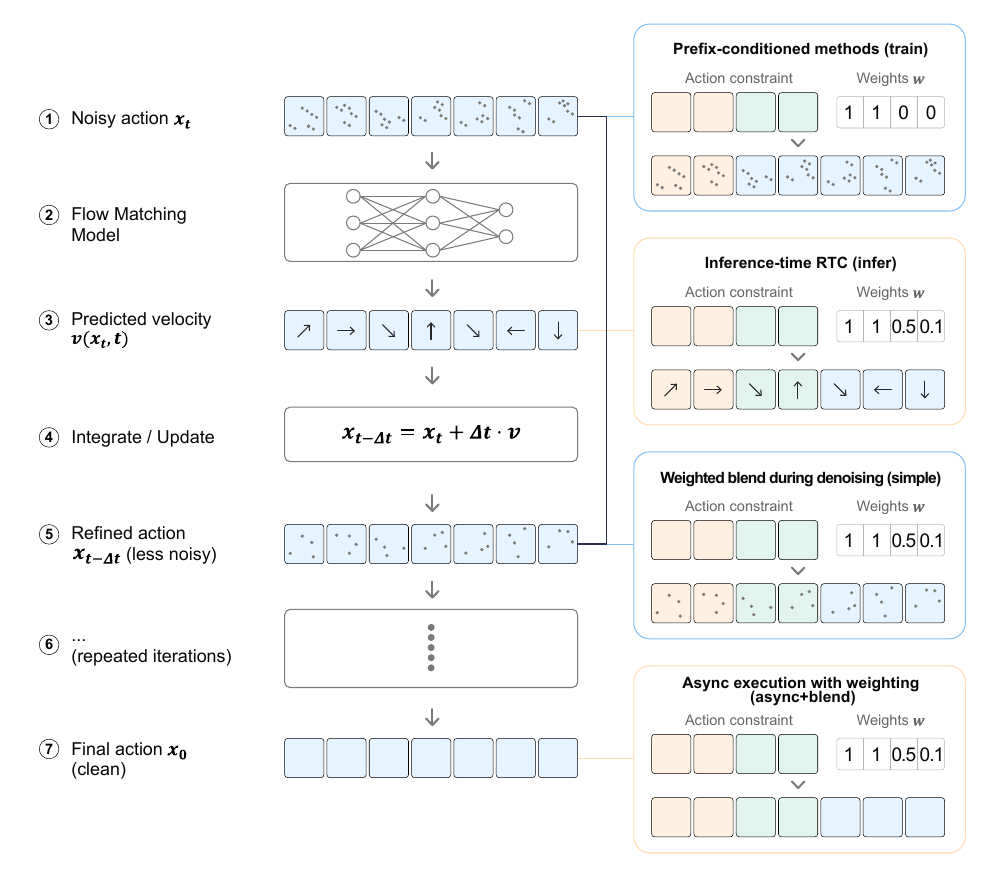}
    \caption{
    Comparison of the four deployment strategies beyond the synchronous
    and pure asynchronous baselines.
    \texttt{async+blend} combines outgoing and incoming action chunks
    post hoc;
    \texttt{simple} blends predicted actions during denoising;
    \texttt{infer} guides the denoising velocity toward consistency with
    the previous chunk; and
    \texttt{train} conditions generation on a committed clean prefix
    learned during training.
    The \texttt{sync} and \texttt{async} baselines are omitted because
    neither applies blending or conditioning.
    }
    \label{fig:method_comparison}
\end{figure*}

We evaluate two execution schedules and four methods for handling transitions between consecutive action chunks. Synchronous (\texttt{sync}) and asynchronous (\texttt{async}) execution specify when inference is triggered and how newly generated chunks are scheduled, without applying any blending. Under asynchronous execution, we further compare four blending methods: post-hoc action weighting (\texttt{async+blend}), weighted blending
during denoising (\texttt{simple}), inference-time RTC (\texttt{infer}), and prefix-conditioned training-time methods (\texttt{train}). Figure~\ref{fig:method_comparison} illustrates and compares these four blending methods. We describe all methods below in increasing order of complexity.

\paragraph{Synchronous Execution (\texttt{sync}).}
\label{sec:sync}
The baseline: the robot halts or continues replaying the last action until inference finishes, then immediately switches to the new chunk. No blending is applied. Latency is fully exposed at every chunk boundary, producing a step discontinuity in commanded velocity.

\paragraph{Pure Asynchronous Execution (\texttt{async}).}
\label{sec:async}
Inference is triggered after executing $s$ frames of the current chunk. When the new chunk arrives, the controller switches at frame $d_\text{est}$ of the new chunk---the frame predicted to correspond to the current robot state. No blending is applied in the overlap region.

This approach eliminates the pause between chunks that occurs in synchronous execution, but provides no smoothness guarantee: the controller performs a hard switch between chunks, which can produce discontinuities if the two chunks disagree.

\paragraph{Asynchronous Execution with Weighting (\texttt{async+blend}).}
\label{sec:asyncblend}
When the new chunk arrives, the controller applies post-hoc weighting between the outgoing and incoming action sequences in the overlap region~\cite{shukor2025smolvlavisionlanguageactionmodelaffordable}. No changes are made to the denoising process or training. This method requires only accurate temporal alignment ($d_\text{est}$) and a simple weighted average, and serves as the minimal blending baseline above pure asynchronous switching.

\paragraph{Weighted Blend During Denoising (\texttt{simple}).}
\label{sec:simple}
During the denoising process that generates chunk $n{+}1$, at each denoising step the predicted actions in the overlap region are blended with the corresponding actions from chunk $n$ (which serve as constraints)~\cite{holobrain2026}. The blending weight at overlap frame $t$ is $w(t) = 1$ for $t \leq d_\text{est}$ and decreases to zero at $t = H - s$, giving a blended prediction $\hat{a}(t) = w(t)\,a_n(t) + (1{-}w(t))\,a_{n+1}(t)$. This is the simplest blending strategy, but introduces a precision--smoothness trade-off: the blended actions are a compromise between the two chunks' predictions, which can reduce task success on precision-critical placements.

\paragraph{Inference-Time RTC (\texttt{infer}).}
\label{sec:infer}
Inference-time RTC~\cite{black2025rtc} modifies the denoising \textit{velocity} (the direction of the denoising trajectory) rather than the generated actions themselves. At each denoising step, the velocity field is adjusted to steer the denoising process toward consistency with chunk $n$'s delay region, using the same weight schedule as above: $w(t) = 1$ for $t \leq d_\text{est}$ and decreasing to zero at $t = H - s$. This provides guidance without directly overwriting the predicted actions, preserving more of the model's original prediction while still reducing inter-chunk disagreement.

\paragraph{Prefix-Conditioned Methods (\texttt{train}).}
\label{sec:train}
Prefix-conditioned methods condition the generation of chunk $n{+}1$ on the committed prefix from chunk $n$ (the delay region). Training-Time RTC~\cite{black2024training} implements this by inserting the first $d_\text{est}$ frames of chunk $n$ as clean conditioning during \textit{training}, teaching the model to naturally continue from a given prefix. Concretely, the weight is $w(t) = 1$ for $t \leq d_\text{est}$ and $w(t) = 0$ beyond---a step function that fully constrains the delay region while leaving the remainder unconstrained, in contrast to the gradually decreasing weights in Section~\ref{sec:simple}. At deployment, the model receives chunk $n$'s delay region as input and generates the remainder of chunk $n{+}1$ conditioned on it.

Legato~\cite{liu2025legato} combines prefix-conditioned training with weighted blending during denoising, merging both mechanisms. SimpleRTC~\cite{holobrain2026} follows a similar design. This family requires model retraining but adds no inference-time overhead.

\section{Experiments}
\label{sec:experiments}

\subsection{Setup}

\textbf{Robot platform.} We deploy all methods on a bimanual end-effector robot controlled at 10\,Hz. The World Action Model predicts chunks of length $H = 24$ frames.

\textbf{Pipeline latency.} We set $d_\text{est} = 8$ frames and $s = 4$ frames, giving an overlap of $H - s = 20$ frames (8-frame delay region + 12-frame remaining overlap). $d_\text{est}$ is set to the median measured end-to-end latency.

\textbf{Tasks.} We evaluate across multiple manipulation tasks spanning a range of precision requirements.

\textbf{Methods.} We compare six strategies: \texttt{sync}, \texttt{async}, \texttt{async+blend}, \texttt{simple}, \texttt{infer}, and \texttt{train}, as described in Section~\ref{sec:methods}.

\subsection{Offline Trajectory Analysis}

Before running on hardware, we perform an offline open-loop evaluation: for each demonstration trajectory in the training set, we sample two observations separated by $s$ frames and feed them independently to the model, obtaining two predicted chunks that simulate consecutive async inference calls. We replay recorded demonstration data through four methods---\texttt{async}, \texttt{simple}, \texttt{infer}, and \texttt{train}---and measure position error in the overlap region. We exclude \texttt{sync} (which has no overlap region) and \texttt{async+blend} (whose post-hoc weighting is applied after inference and does not affect the chunk predictions themselves). We report both mean and maximum error: while mean error reflects overall trajectory quality, a single large discontinuity (captured by the max metric) can push the robot into an anomalous state, making max error task-critical.

We decompose the overlap into two regions:

\begin{enumerate}
  \item \textbf{Delay-region error ($\Delta_\text{delay}$)}: L2 distance between the two chunks' predictions in the $d_\text{est}$-frame delay region. Ideally this should approach zero---a method that strongly constrains this region guarantees smooth execution when $d = d_\text{est}$.
  \item \textbf{Remaining-overlap error ($\Delta_\text{rem}$)}: L2 distance in the overlap frames beyond $d_\text{est}$. When $d > d_\text{est}$ or chunk alignment is imprecise, these frames are the ones actually executed at the chunk boundary, and their error directly determines the magnitude of the position jump.
\end{enumerate}

\begin{figure}[!htbp]
  \centering
  \includegraphics[width=\linewidth]{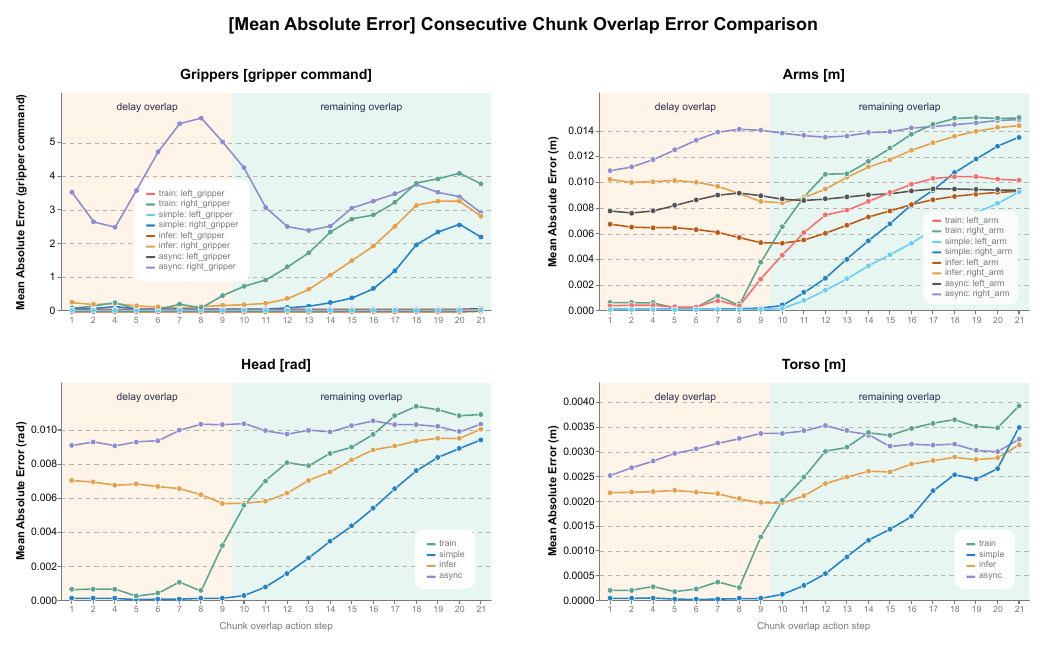}
  \caption{Mean absolute error (MAE) as a function of overlap action step, shown separately for four components: arm (position error, m), torso (position error, m), head (joint angle error, rad), and gripper (0--100). The delay region (steps 1--8) is highlighted in yellow; the remaining overlap (steps 9--20) in green. Lower error indicates stronger action constraint in the overlap region.}
  \label{fig:offline_mae}
\end{figure}

\begin{figure}[!htbp]
  \centering
  \includegraphics[width=\linewidth]{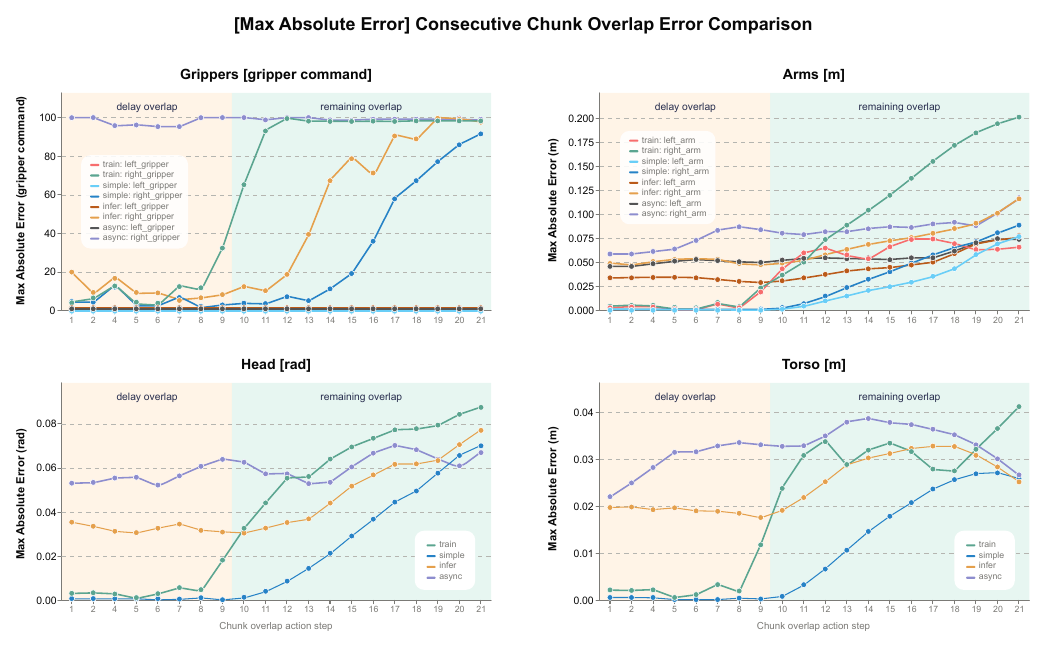}
  \caption{Maximum absolute error as a function of overlap action step, shown separately for arm, torso, head, and gripper. Same color coding and units as Figure~\ref{fig:offline_mae}. Lower error indicates stronger action constraint in the overlap region.}
  \label{fig:offline_max}
\end{figure}

\textbf{Results.} Figures~\ref{fig:offline_mae} and~\ref{fig:offline_max} show per-joint MAE and max absolute error as a function of overlap action step, with the delay region (steps 1--8) highlighted in yellow and the remaining overlap (steps 9--20) in green.

\textbf{Key observations:}

\begin{enumerate}
  \item \textbf{\texttt{infer} fails to constrain the delay region.} Across all joints, \texttt{infer} shows substantially higher MAE and max error in the delay region compared to \texttt{simple} and \texttt{train}. Unlike action-constraining methods, velocity-level guidance steers the denoising trajectory without enforcing hard constraints on the output actions, and therefore cannot guarantee action continuity at chunk boundaries---position jumps remain unavoidable.

  \item \textbf{\texttt{async} has the worst errors overall.} Without any blending, \texttt{async} produces the highest errors in both regions and both metrics, confirming that unblended async switching is not viable without temporal alignment.

  \item \textbf{\texttt{simple} and \texttt{train} both constrain the delay region effectively.} Both methods maintain near-zero error in the delay region (steps 1--8) across all joints. In the remaining overlap, \texttt{simple} stays consistently lower as its aggressive interpolation continues to enforce proximity between chunks, while \texttt{train}'s error grows since it only conditions on the delay region and preserves the model's original predictions beyond it.

  \item \textbf{The delay region boundary is the critical threshold.} The transition from yellow to green region marks a clear inflection point: all methods show increasing error in the remaining overlap, and the gap between methods widens. This underscores the importance of accurate $d_\text{est}$---if the true delay $d$ exceeds $d_\text{est}$, execution falls into the higher-error remaining overlap region.
\end{enumerate}

\subsection{Online Robot Evaluation}

\begin{figure}[htbp]
  \centering
  \includegraphics[width=\linewidth]{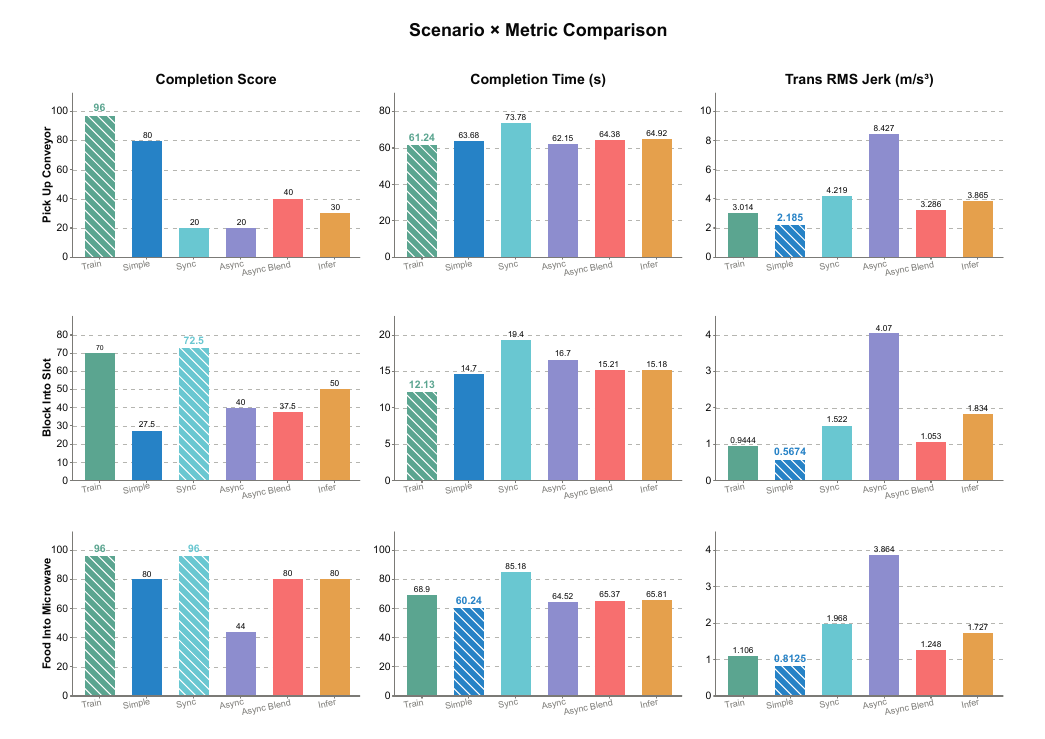}
  \caption{Online evaluation results (completion score, completion time, and translational RMS jerk) across three tasks and six methods.}
  \label{fig:online_results}
\end{figure}

We evaluate on three tasks that stress different aspects of asynchronous deployment: \textbf{Pick Up Conveyor} (dynamic task---moving target requiring fast closed-loop response), \textbf{Block Into Slot} (fine manipulation---precision-critical placement), and \textbf{Food Into Microwave} (long-horizon task---extended sequence with lower per-step precision demands). Each method--task combination is evaluated over 5 trials; reported completion scores, completion times, and jerk values are averaged across trials. Results are shown in Figure~\ref{fig:online_results}.

\textbf{Scoring rubric.} Completion score is computed from sub-task milestones, each worth a fixed number of points:
\begin{itemize}
  \item \textit{Pick Up Conveyor} (5 objects, max 100): picking up each object (+10), placing it into the bin (+10).
  \item \textit{Block Into Slot} (max 100): locating the target block (+25), grasping the block (+25, with a second attempt allowed; a failed first attempt deducts 12.5), moving the block above the target slot (+25), inserting the block into the slot (+25).
  \item \textit{Food Into Microwave} (max 100): pressing the microwave switch (+20), opening the microwave door (+20), grasping the bowl of food (+20), placing the bowl inside the microwave (+20), closing the microwave door (+20).
\end{itemize}

\textbf{Dynamic task: Pick Up Conveyor.}
\texttt{sync} achieves a completion score of only 20, failing to track the moving target because the robot is completely unresponsive during the inference window. \texttt{async} also scores 20 despite being responsive---the hard chunk-switch without any blending produces the highest jerk among all methods, severe enough to prevent successful picking. \texttt{async+blend} (score 40) demonstrates that accurate temporal alignment combined with even the simplest interpolation is the minimal prerequisite: only once chunks are properly aligned and blended does the baseline begin to yield passable trajectories. Among methods with stronger blending, \texttt{train} achieves the highest score (96) with competitive completion time (61.24\,s) and low jerk. \texttt{simple} also performs well (score 80), benefiting from its aggressive action smoothing in a task that is not precision-critical. \texttt{infer} scores 30, limited by its larger effective $d_\text{est}$ due to additional inference overhead.

\textbf{Fine manipulation task: Block Into Slot.}
\texttt{simple}'s aggressive action interpolation degrades placement precision (score 27.5), confirming that blending during denoising introduces a precision--smoothness trade-off on contact-critical tasks. \texttt{sync} (score 72.5) and \texttt{train} (score 70) both achieve high completion scores; \texttt{train} additionally completes the task faster (12.13\,s vs.\ 19.4\,s for \texttt{sync}) and with lower jerk, as prefix conditioning reduces inter-chunk disagreement without sacrificing prediction accuracy. \texttt{async} alone produces the highest jerk with moderate completion (score 40); adding blend reduces jerk but does not recover precision (score 37.5), further motivating methods that constrain inter-chunk consistency at training time.

\textbf{Long-horizon task: Food Into Microwave.}
Per-step precision demands are low, so both \texttt{train} and \texttt{sync} achieve high completion scores (96). However, \texttt{sync} incurs a pause at every chunk boundary, accumulating into a 85.18\,s completion time---substantially longer than \texttt{train} (68.9\,s) and all asynchronous methods ($\sim$60--66\,s). \texttt{simple} completes fastest (60.24\,s) with the lowest jerk and a score of 80. \texttt{async} alone drops to 44 with high jerk; \texttt{async+blend} reaches 80 with reduced jerk---a passable outcome that nonetheless leaves room for the precision and smoothness gains offered by \texttt{train}.

\textbf{Cross-task summary.}
Across all tasks (Figure~\ref{fig:online_results}), two engineering baselines stand out. \texttt{async} exposes the raw cost of no inter-chunk reconciliation: high jerk and degraded completion. \texttt{async+blend}, with only accurate temporal alignment and a straightforward weighting step, establishes a workable floor---adequate for low-precision scenarios but insufficient for fine manipulation or optimal smoothness. This baseline motivates the more sophisticated methods: \texttt{simple} pushes smoothness further at a precision cost, \texttt{infer} steers the denoising trajectory at the expense of a larger $d_\text{est}$, and \texttt{train} achieves the best overall trade-off by learning to produce consistent continuations at training time. \texttt{sync} remains competitive on non-dynamic tasks but is universally the slowest and fails on the dynamic scenario.

\begin{figure}[!htbp]
  \centering
  \includegraphics[width=0.7\linewidth]{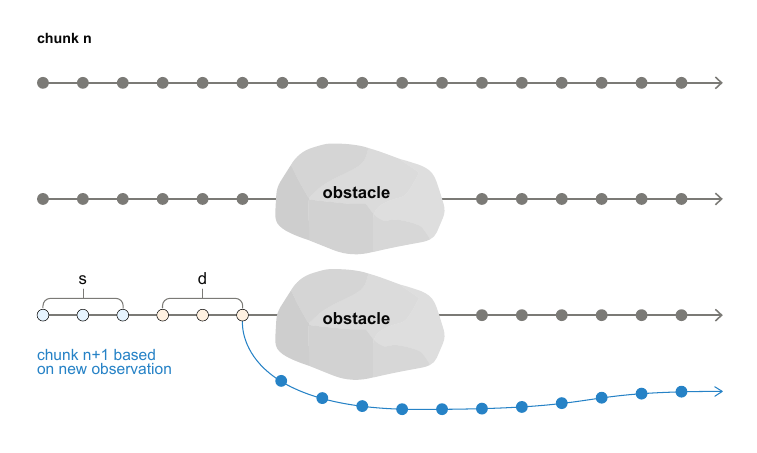}
  \caption{When an obstacle appears unexpectedly, chunk $n{+}1$ must deviate substantially from the prior chunk's trajectory. In this case, the prefix constraint becomes harmful, and none of the blending strategies discussed here are equipped to handle abrupt environmental changes.}
  \label{fig:obstacle}
\end{figure}

\section{Discussion and Conclusion}

\textbf{Discussion.}
Reliable async execution---accurate temporal alignment---is the prerequisite for any blending strategy. Without it, trajectory jitter is large; once the foundation is solid, \texttt{async+blend} already yields acceptable trajectories on low-precision tasks. Among blending methods, direct action weighting (\texttt{simple}) achieves the best smoothness but compromises precision on fine manipulation, as averaging two divergent predictions produces trajectories that satisfy neither. Prefix-conditioned methods (\texttt{train}) avoid this by learning consistent continuations at training time, achieving the best overall results across all three tasks. Velocity-guided inference (\texttt{infer}), despite its theoretical appeal, fails to adequately constrain the delay region on our platform---offline analysis confirms substantially higher $\Delta_\text{delay}$ compared to action-constraining methods.

Finally, all methods here assume the prior chunk provides useful guidance; this breaks down when the environment changes abruptly. As illustrated in Figure~\ref{fig:obstacle}, a suddenly appearing obstacle drives chunk $n{+}1$ away from the prior trajectory, making the prefix constraint harmful rather than helpful. How to generate smooth yet reactive trajectories under such conditions is an open problem we plan to explore.

\textbf{Conclusion.}
We compared six asynchronous deployment strategies for World Action Models across fine manipulation, dynamic, and long-horizon tasks. Engineering quality---specifically temporal alignment---is the necessary foundation. Given proper alignment, direct action weighting provides a smooth baseline; prefix-conditioned methods achieve the best overall precision--smoothness balance; and velocity-guided inference cannot guarantee action continuity at chunk boundaries on our platform.

\section*{Author Contributions}
Mengchen Cai, Jiangfeng Liu, and Yinze Rong contributed to this work. Mengchen Cai initiated the project; Jiangfeng Liu served as the project lead.

\section*{Acknowledgments}
The authors thank Prof. Jun Zhu for his support of this work.

\bibliographystyle{unsrt}
\bibliography{main}

\end{document}